\documentclass[conference]{IEEEtran}
\IEEEoverridecommandlockouts
\usepackage{cite}

\usepackage{soul}
\usepackage{pdfpages}
\usepackage{color}
\usepackage{xcolor}
\usepackage{url}
\usepackage{listings}   
\usepackage{amsfonts}
\usepackage{nicefrac}
\usepackage{microtype}
\usepackage{amsmath}
\usepackage{graphicx}
\usepackage{subfigure}
\usepackage{algorithm}
\usepackage{algpseudocode}
\usepackage{multirow}
\usepackage{mathtools}
\usepackage{balance}
\usepackage{bm}
\usepackage{dsfont}
\usepackage{accents}
\usepackage{booktabs}

\usepackage{matharticle}
\usepackage{algpseudocode}

\usepackage{paralist}
\usepackage{todonotes}
\usepackage{xspace}
\usepackage{booktabs} 
\usepackage[utf8]{inputenc} 
\usepackage[T1]{fontenc}    
\usepackage{hyperref}       
\usepackage{url}            
\usepackage{amsfonts}       
\usepackage{nicefrac}       
\usepackage{microtype}      
\usepackage{comment}

\algblock{ParFor}{EndParFor}
\algnewcommand\algorithmicparfor{\textbf{parfor}}
\algnewcommand\algorithmicpardo{\textbf{do}}
\algnewcommand\algorithmicendparfor{\textbf{end\ parfor}}
\algrenewtext{ParFor}[1]{\algorithmicparfor\ #1\ \algorithmicpardo}
\algrenewtext{EndParFor}{\algorithmicendparfor}

\newcommand{\EGAN}{\textbf{E-GAN}\xspace}
\newcommand{\DCGAN}{\textbf{GAN-BCE}\xspace}
\newcommand{\SEGAN}{\textbf{Mustangs}\xspace}
\newcommand{\SCoevGANmm}{\textbf{Lip-BCE}\xspace}
\newcommand{\SCoevGANls}{\textbf{Lip-MSE}\xspace}
\newcommand{\SCoevGANh}{\textbf{Lip-HEU}\xspace}

\usepackage{mwe}
\usepackage{graphicx}

\usepackage{color, colortbl}

\begin{document}

\title{Fostering Diversity in Spatial Evolutionary Generative Adversarial Networks}

\author{\IEEEauthorblockN{Jamal Toutouh}
\IEEEauthorblockA{\textit{Massachusetts Institute of Technology} \\
Cambridge, MA, USA \\
toutouh@mit.edu}
\and
\IEEEauthorblockN{Erik Hemberg}
\IEEEauthorblockA{\textit{Massachusetts Institute of Technology} \\
Cambridge, MA, USA \\
hembergerik@csail.mit.edu}
\and
\IEEEauthorblockN{Una-May O'Reilly}
\IEEEauthorblockA{\textit{Massachusetts Institute of Technology} \\
Cambridge, MA, USA \\
unamay@csail.mit.edu}
}

\maketitle

\begin{abstract}
Generative adversary networks (GANs) suffer from training pathologies such as instability and mode collapse, which mainly arise from a lack of diversity in their adversarial interactions. Co-evolutionary GAN (CoE-GAN) training algorithms have shown to be resilient to these pathologies. This article introduces Mustangs, a spatially distributed CoE-GAN, which fosters diversity by using different loss functions during the training. Experimental analysis on MNIST and CelebA demonstrated that Mustangs trains statistically more accurate generators.
\end{abstract}


\setlength{\intextsep}{3pt}

\vspace{-0.5cm}
\section{Introduction}
\vspace{-0.2cm}

Generative adversarial networks (GANs) have emerged as a powerful
machine learning paradigm for the task of estimating
a distribution function underlying a given set of samples~\cite{goodfellow2014generative}. 
The successes of GANs in generating realistic, complex, multivariate distributions
motivated a growing body of applications, such as realistic image
generation~\cite{gan2017triangle}, video
prediction~\cite{liang2017dual}, and text to image
synthesis~\cite{reed2016learning}.

A GAN consists of two neural networks (NN), a generator and a
discriminator, which apply adversarial learning to train their weights against each other (formulated as a minmax optimization problem). 
The discriminator is trained to correctly discern the
``natural/real'' samples from ``artificial/fake'' samples produced by
the generator. The generator is trained
to transform a random input into samples that fool the
discriminator. 
GANs are notoriously hard to train, frequently showing pathologies such as mode collapse or vanishing gradients~\cite{wang2019evolutionary,schmiedlechner2018towards}. 

Co-evolutionary training (CoE) has shown to be resilient to the GAN training degenerative behaviors by evolving two populations of NNs (one of generators against one of discriminators) towards convergence while keeping genome diversity~\cite{schmiedlechner2018towards}. 
This article summarizes our previous research~\cite{mustangs}, in which we proposed a CoE GAN training method named MUtation SpaTial gANs (\SEGAN). 
\SEGAN combines the ideas of two successful GAN training approaches that foster diversity by applying different strategies: Evolutionary GAN (E-GAN), which injects diversity by applying three different NN mutations~\cite{wang2019evolutionary}, and Lipizzaner, which applies a spatially distributed cellular CoE traiining~\cite{schmiedlechner2018lipizzaner}. 
The main aim of this research was to show that combining ideas from both training methods improve diversity, and therefore, is better than either one of them.
%

\section{MUtation SpaTial gANs training}
\vspace{-0.2cm}

\SEGAN applies a spatially distributed CoE to train GANs. It evolves two populations, $P_u=\{u_1, \ldots, u_T\}$ a population of generators and $P_v=\{v_1, \ldots, v_T\}$ a population of discriminators to create diversity in genomes spaces. 
Both populations ($P_u$ and $P_v$) are distributed on the cells of a two dimensional toroidal grid~\cite{schmiedlechner2018towards}. 
The fitness $\calL$ of each generator $u_i\in P_u$ and discriminator $v_j \in P_v$ are assessed according to their interactions with a set of discriminators from $P_v$ and  generators from $P_u$, respectively. 
The fittest individuals are used to generate the new of individuals (generators and discriminators) by applying mutation. 
The new individuals replace the ones in the current population if they perform better (better fitness) to produce the next generation.

Instead of applying the CoE in an \emph{all-vs-all} flavor, the cell's \emph{neighborhood} defines the subset of individuals (sub-populations) of $P_u$ and $P_v$ to interact with and it is specified by its size $s_{n}$.
In our study, we use a five-cell neighborhood, i.e, one center and four adjacent cells. 
%
In \SEGAN, each neighborhood (sub-population) performs an instance of the CoE to update its center cell with the fittest NN after each training epoch. 
Besides, the fittest individuals are sent to the neighborhood cells to update the sub-populations of the grid.

\begin{figure}[!h]
\centering
\vspace{-0.1cm}
\includegraphics[width=0.8\linewidth, trim={2cm 5.2cm 5cm 1.6cm},clip]{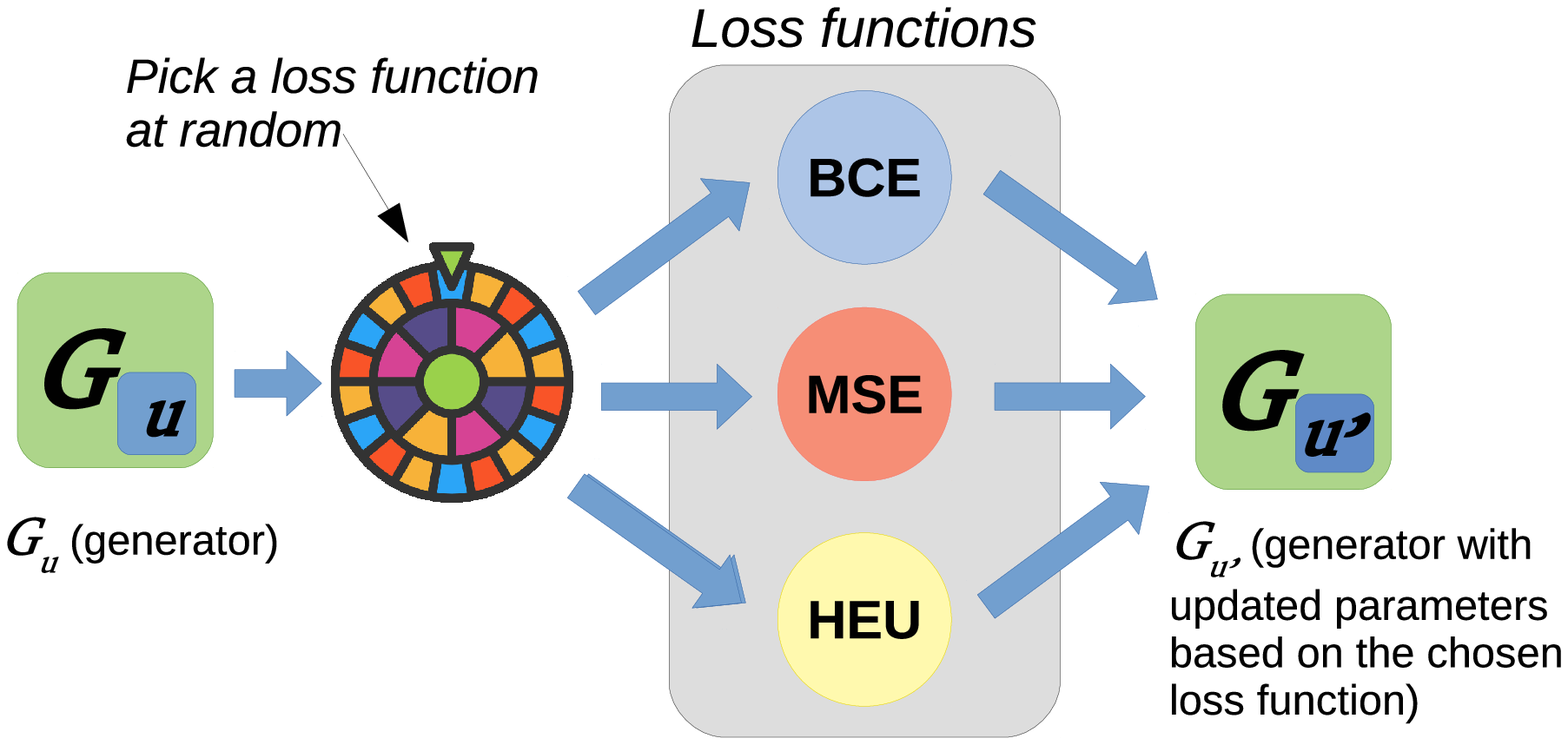}
\vspace{-0.2cm}
\caption{\small Mutation used in \SEGAN.}
\label{fig:segan-mutation}
\end{figure}

At the time of writing the article, CoE applied the mutation based on stochastic gradient descent (SGD) to create the offspring by minimizing only one objective/loss function, which generally attempts to minimize the distance between the generated fake data and real data distributions according to a given measure~\cite{schmiedlechner2018lipizzaner}. 
Instead, 
\SEGAN generates the offspring by an applying SGD-based mutation that randomly picks a giving training objective function (see Fig.\ref{fig:segan-mutation}). 
\SEGAN applies mutations introduced by E-GAN~\cite{wang2019evolutionary}: \emph{Minmax mutation} (BCE), \emph{Least-square mutation} (LSE), and \emph{Heuristc mutation} (HEU).

\section{Experimental analysis}
\vspace{-0.2cm}

This section summarizes the main outcomes of the empirical analysis shown in our previous work~\cite{mustangs}. 
\SEGAN was evaluated on two common image data sets: MNIST and CelebA. 
The experiments took into account: \DCGAN, a standard GAN which uses BCE objective; E-GAN; a spatial CoE that applies each one of the objective functions, \SCoevGANmm, \SCoevGANls, and \SCoevGANh; and \SEGAN. All evaluated methods used the same computational budget.

The Frechet inception distance (FID) was evaluated to asses the accuracy of the generated fake data~\cite{heusel2017gans} and the total variation distance (TVD) for diversity. 

Table~\ref{tab:fid-results} shows the best FID values on MNIST. \SEGAN has the lowest (best) median. All the spatially distributed methods are better than \EGAN and \DCGAN. The results indicate that \SEGAN is robust to the varying performance of the individual loss functions (lowest deviation). This helps to strengthen the idea that diversity, both in genome and mutation space, provides robust GAN training. 
A ranksum test with Holm correction confirms that the difference between
\SEGAN and the other methods is significant at confidence levels of
$\alpha<$0.01.

\begin{table}[!h]
	\centering
	\scriptsize
	\caption{\small FID MNIST results(Low FID indicates good performance)}
	\vspace{-0.2cm}
	\label{tab:fid-results}
	\begin{tabular}{lrrrr}
	    \toprule
		\textbf{Algorithm} & \textbf{Mean} & \textbf{Std\%} & \textbf{Median} & \textbf{IQR}  \\ \hline
\SEGAN & \textit{42.235} & 12.863\% & \textit{43.181} &  7.586 \\ 
\SCoevGANmm & 48.958 & 20.080\% & 46.068 &  4.663 \\ 
\SCoevGANls & 371.603 & 20.108\% & 381.768 &  104.625 \\ 
\SCoevGANh & 52.525 & 17.230\% & 52.732 &  9.767 \\ 
\EGAN & 466.111 & 10.312\% & 481.610 &  69.329 \\ 
\DCGAN & 457.723 & 2.648\% & 459.629 &  17.865 \\ 
		\bottomrule
	\end{tabular}
	\vspace{-.cm}
\end{table}

Table~\ref{tab:tvd} summarizes TVD results on MNIST. 
The methods that provide genome diversity generate more diverse data samples than the other two analyzed methods.
The three algorithms with the lowest (best) FID score (\SEGAN, \SCoevGANmm, and \SCoevGANh) also provide the lowest (best) TVD values. The best TVD result is obtained by \SCoevGANh. 

\begin{table}[!h]
\setlength{\tabcolsep}{3pt} 
    \renewcommand{\arraystretch}{1}
	\centering
	\scriptsize
	\caption{\small MNIST TVD results (Low TVD indicates more diversity)}
	\vspace{-0.1cm}
	\label{tab:tvd}
	\begin{tabular}{lrrrrrr}
	    \toprule
		\textbf{Alg.} & \SEGAN & \SCoevGANmm & \SCoevGANh & \SCoevGANls & \EGAN & \DCGAN \\ \hline
		\textbf{TVD} & 0.180 & 0.171 & \textit{0.115} & 0.365 & 0.534 & 0.516 \\
		\bottomrule
	\end{tabular}
	\vspace{0.cm}
\end{table}

\begin{figure}[!h]
\setlength{\tabcolsep}{1pt} 
\renewcommand{\arraystretch}{0.8} 
		\begin{tabular}{cccc}

		\includegraphics[width=0.12\textwidth, trim={0 63mm 0 0},clip]{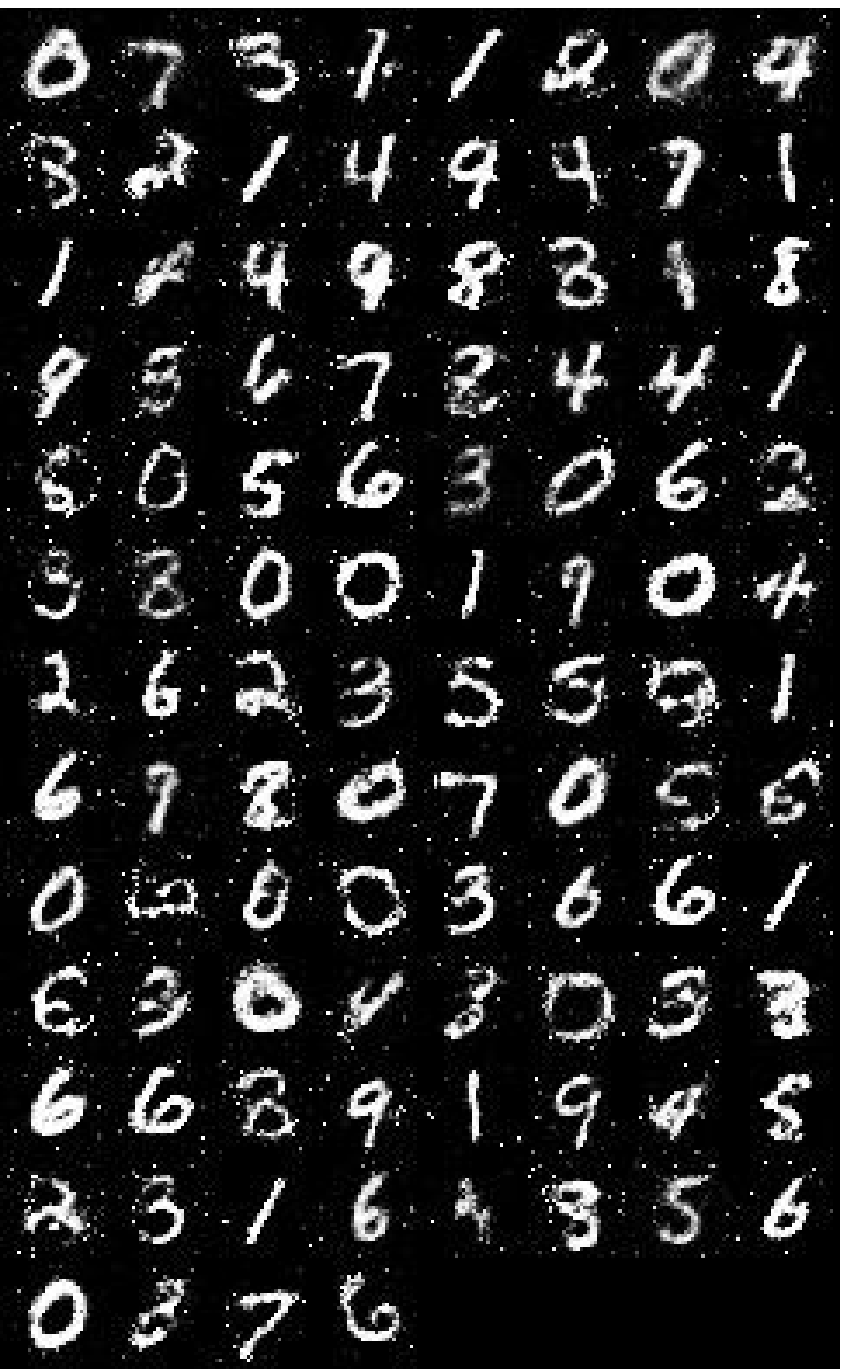} 
		& \includegraphics[width=0.12\textwidth, trim={0 63mm 0 0},clip]{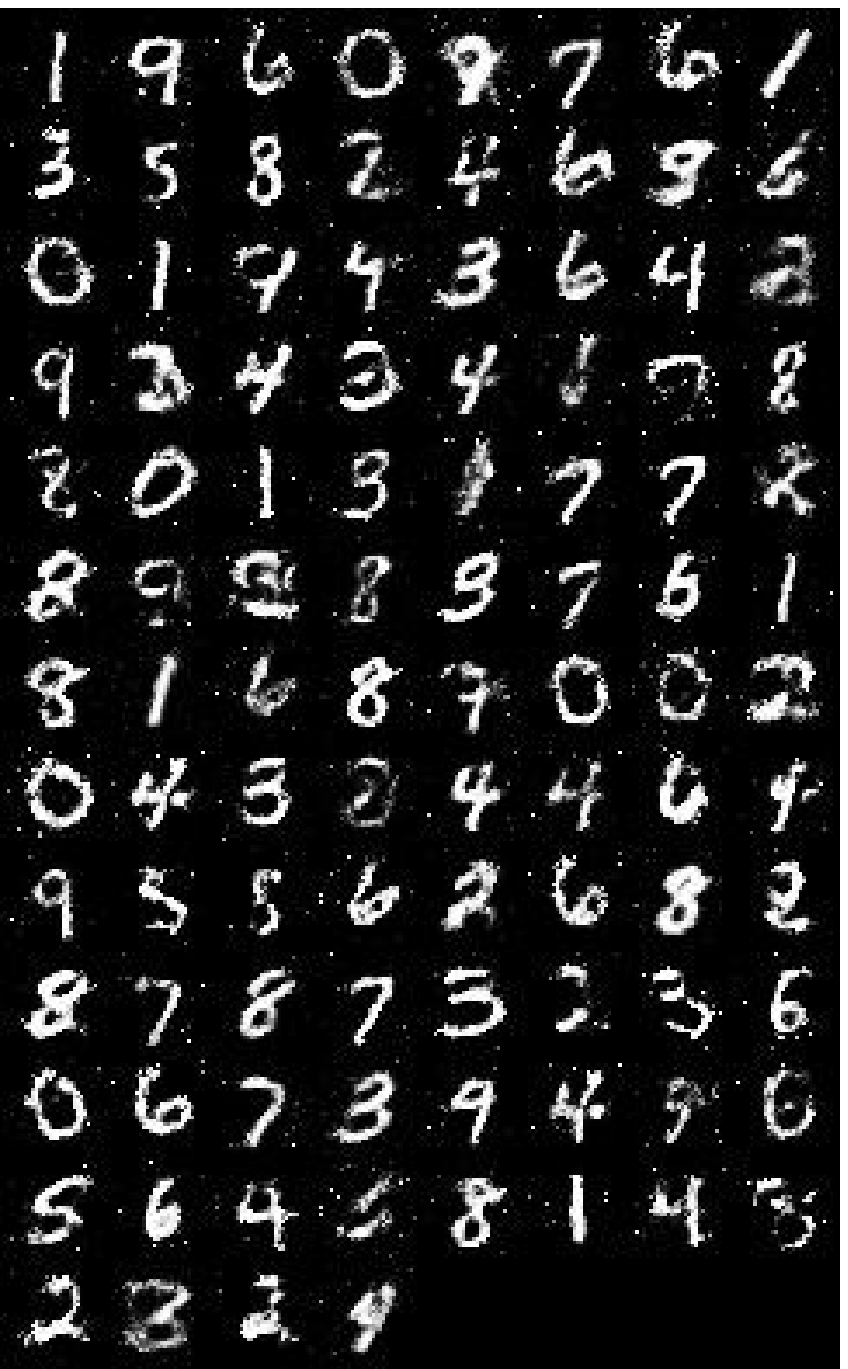} 
		& \includegraphics[width=0.12\textwidth, trim={0 63mm 0 0},clip]{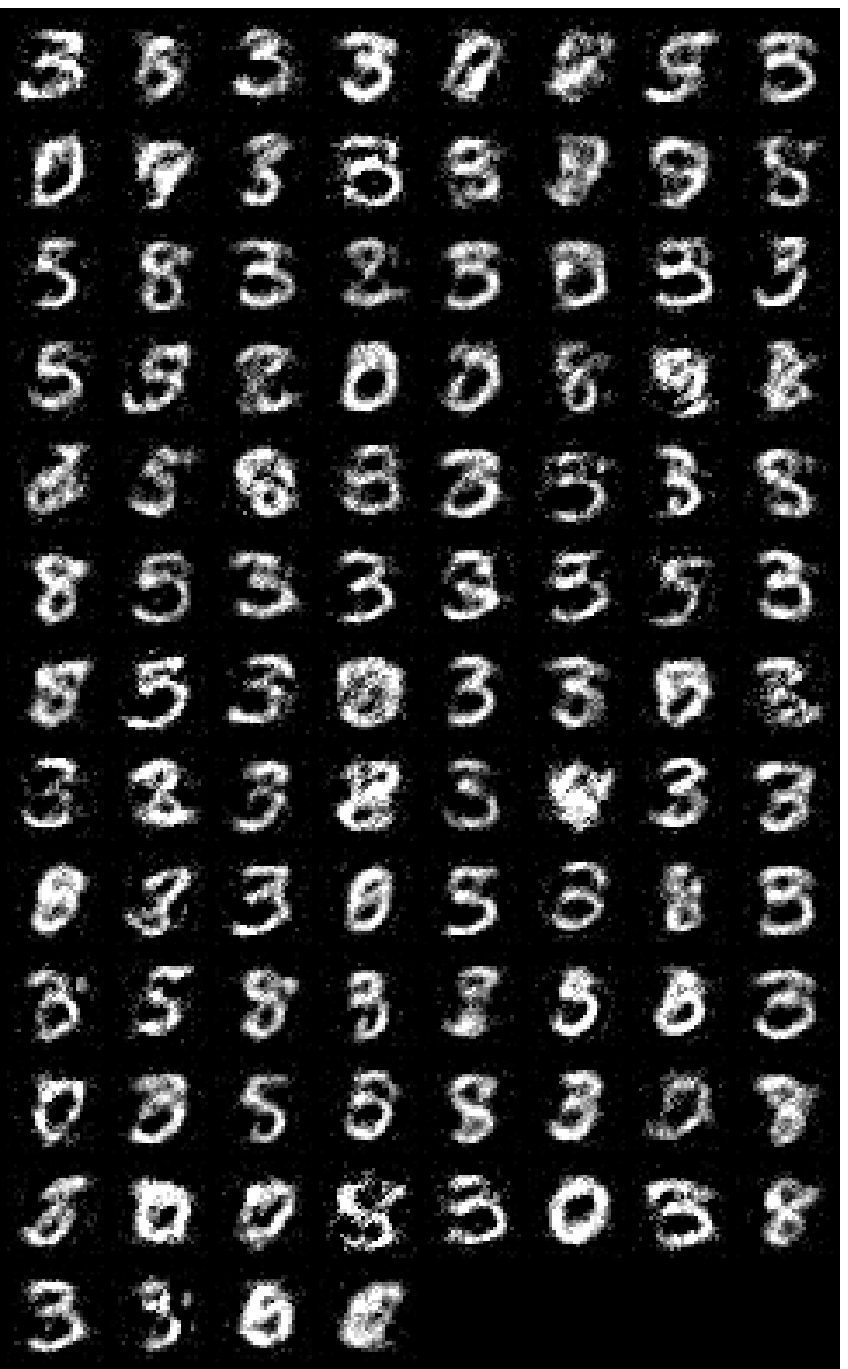} & \includegraphics[width=0.12\textwidth, trim={0 63mm 0 0},clip]{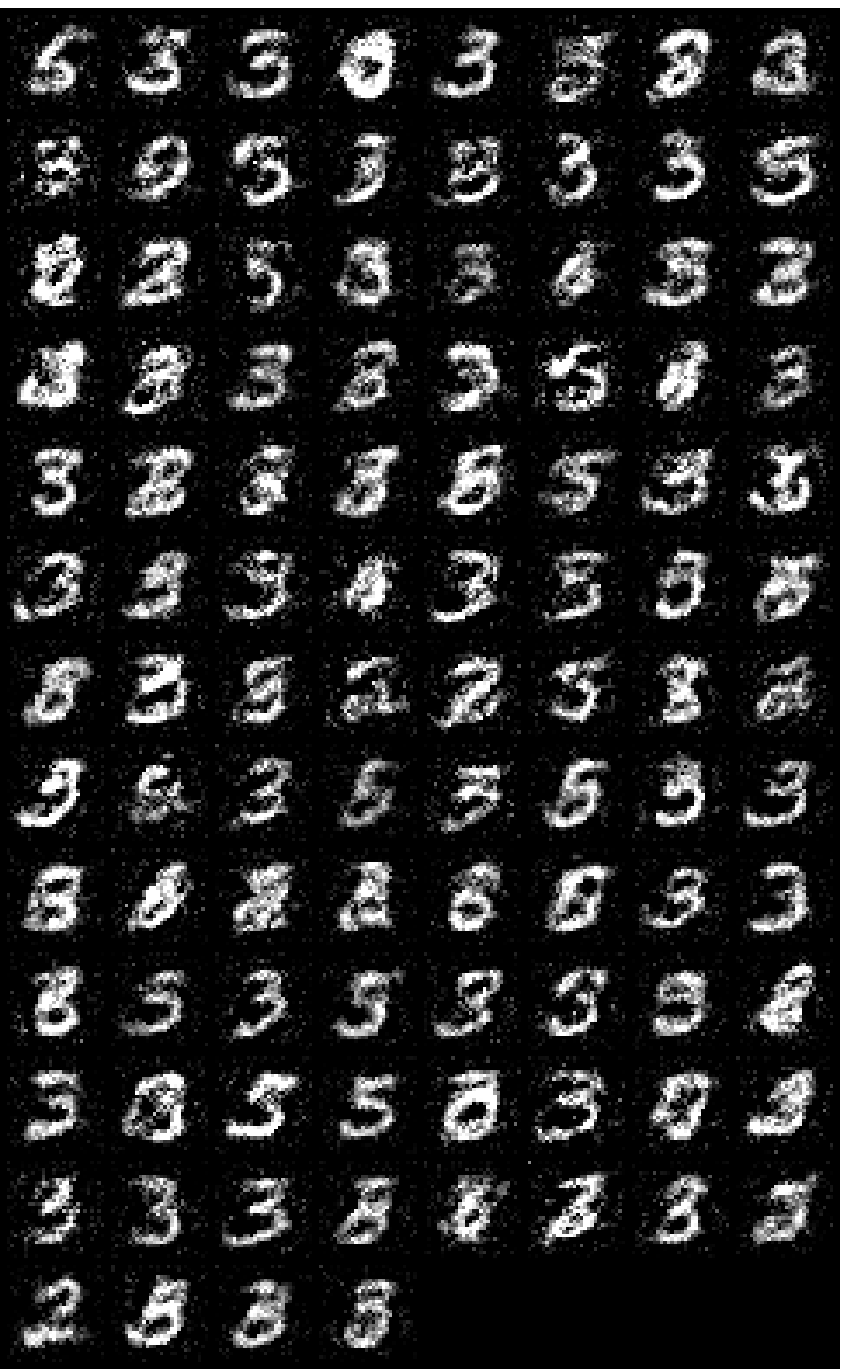} \\
		a) \SEGAN & b) \SCoevGANmm & c) \EGAN & d) \DCGAN
	\end{tabular}		
	\caption{Sequence of samples generated of MNIST dataset.}
	\label{fig:mnist}
	\vspace{-0.0cm}
\end{figure}

Figure~\ref{fig:mnist} illustrates how spatially distributed CoE algorithms produce robust generators that provide with accurate MNIST samples across all the classes. 

Table~\ref{tab:fid-results-celeba} summarizes FID the results on CelebA. 
\SEGAN provides the lowest median FID. 
\SCoevGANmm and \SCoevGANh provide median and mean FIDs close to the \SEGAN ones.  
However, \SEGAN is the most robust (see deviations).

\begin{table}[!h]
	\centering
	\scriptsize
	\caption{\small FID CelebA results (Lower is better)}
	\vspace{-0.2cm}
	\label{tab:fid-results-celeba}
	\begin{tabular}{lrrrr}
	    \toprule
		\textbf{Algorithm} & \textbf{Mean} & \textbf{Std\%} & \textbf{Median} & \textbf{IQR}  \\ \hline
\SEGAN & 36.148 & 0.571\% & 36.087 &  0.237 \\ 
\SCoevGANmm & 36.250 & 5.676\% & 36.385 &  2.858 \\ 
\SCoevGANls & 158.740 & 40.226\% & 160.642 &  47.216 \\ 
\SCoevGANh & 37.872 & 5.751\% & 37.681 &  2.455 \\ 
		\bottomrule
	\end{tabular}
	\vspace{-.4cm}
\end{table}

\begin{figure}[!h]
\setlength{\tabcolsep}{4pt} 
\renewcommand{\arraystretch}{0.8} 
\centering
	\begin{tabular}{cc}
	\includegraphics[width=0.4\linewidth, trim={0 140mm 0 0},clip]{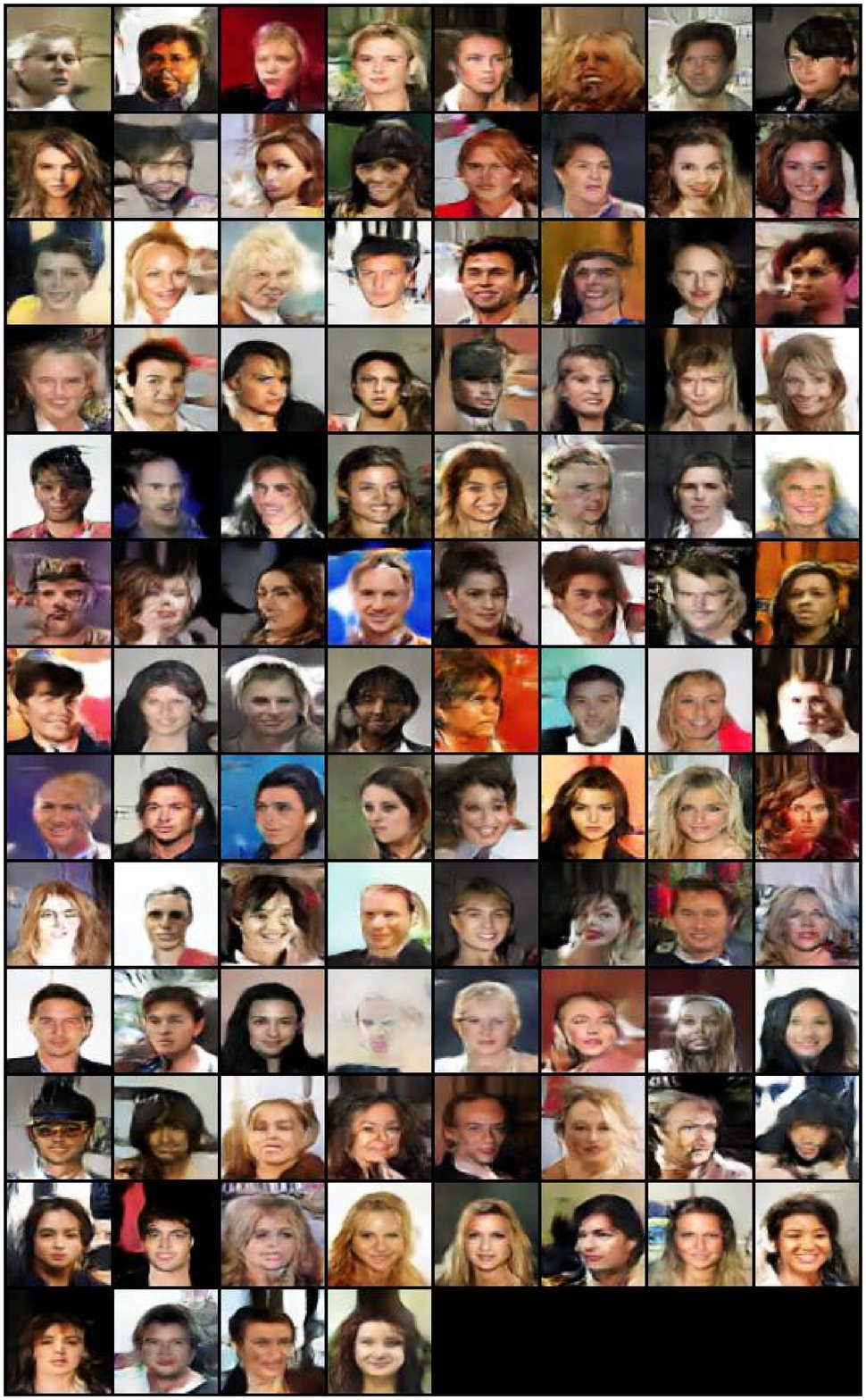} &
		\includegraphics[width=0.4\linewidth, trim={0 140mm 0 0},clip]{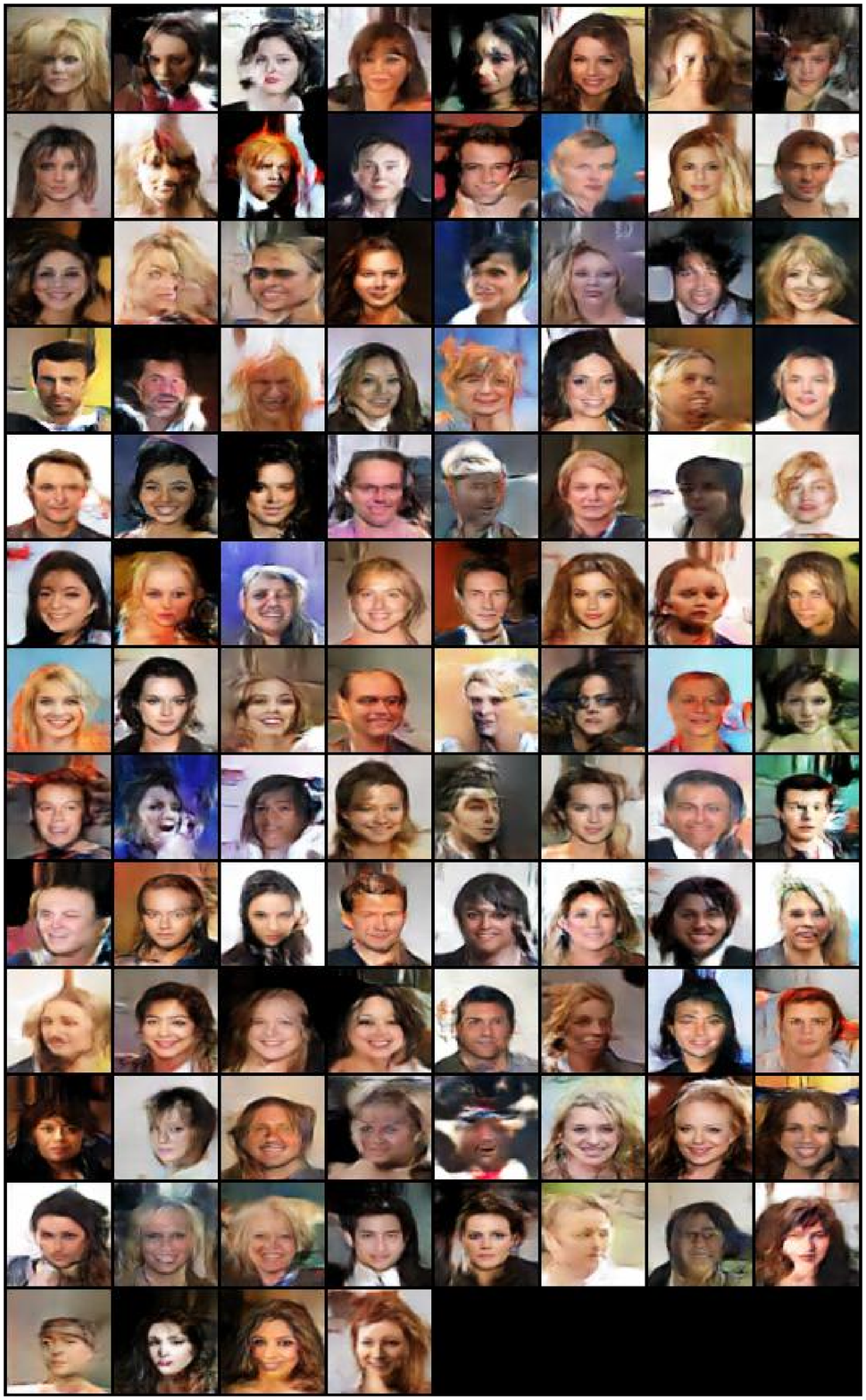} \\
		(a) \SEGAN & (b) \SCoevGANmm 
	\end{tabular}	
	\caption{Sequence of samples generated of CelebA. }
	\label{fig:celeba-samples}
\end{figure}

Figure~\ref{fig:celeba-samples} illustrates a sequence of samples generated by the best generators in terms of FID score of the most competitive two training methods, i.e., \SEGAN and \SCoevGANmm. 
As it can be seen in these two sets images generated, both methods present similar capacity of generating human faces. 

\section{Conclusions and Future Work}

The research carried out showed that GAN training can be improved by boosting diversity. Mustangs tested on the MNIST and CelebA datasets showed the best accuracy, robustness, and high diversity in label space. This research was the root of other published studies~\cite{toutouh2020data,toutouh2020analyzing}.

\bibliographystyle{IEEEtran}
\bibliography{bibliography}

\end{document}